%% file: paper.tex
\pdfoutput=1

\documentclass[11pt]{article}

\usepackage[preprint]{acl}

\usepackage{times}
\usepackage{latexsym}

\usepackage[T1]{fontenc}

\usepackage[utf8]{inputenc}

\usepackage{microtype}

\usepackage{inconsolata}

\usepackage{graphicx}

\usepackage{xspace}
\usepackage{amsmath}
\usepackage{booktabs}
\usepackage{multirow}
\usepackage{arydshln}
\usepackage{cleveref}
\usepackage{subcaption}

\newcommand{\base}{$\mathtt{Llama\text{-}3.1\text{-}8B\text{-}Instruct}$\xspace}
\newcommand{\gradient}{$\mathtt{Llama\text{-}3\text{-}8B\text{-}Instruct\text{-}Gradient\text{-}1048k}$\xspace}
\newcommand{\prolong}{$\mathtt{Llama\text{-}3\text{-}8B\text{-}ProLong\text{-}512k\text{-}Instruct}$\xspace}

\newcommand{\basetext}{Llama-3.1-8B-Instruct\xspace}
\newcommand{\gradienttext}{Llama-3-8B-Instruct-Gradient-1048k\xspace}
\newcommand{\prolongtext}{Llama-3-8B-ProLong-512k-Instruct\xspace}

\newcommand{\modelIbasemulti}{$\mathtt{UltraLong\text{-}8B\text{-}512k\text{-}1M}$}
\newcommand{\modelIIbasemulti}{$\mathtt{UltraLong\text{-}8B\text{-}1M\text{-}2M}$}

\newcommand{\modelIbase}{$\mathtt{UltraLong\text{-}8B\text{-}1M}$}
\newcommand{\modelIIbase}{$\mathtt{UltraLong\text{-}8B\text{-}2M}$}
\newcommand{\modelI}{$\mathtt{UltraLong\text{-}8B\text{-}1M\text{-}Instruct}$}
\newcommand{\modelII}{$\mathtt{UltraLong\text{-}8B\text{-}2M\text{-}Instruct}$}
\newcommand{\modelIV}{$\mathtt{UltraLong\text{-}8B\text{-}4M\text{-}Instruct}$}

\title{From 128K to 4M: Efficient Training of Ultra-Long Context \\ Large Language Models}

\author{
 \textbf{Chejian Xu\thanks{Work done during an internship at NVIDIA. \; \textsuperscript{1}UIUC, \textsuperscript{2}NVIDIA. \; 
  †Correspondence to: Chejian Xu  <chejian2@illinois.edu>, Wei Ping <wping@nvidia.com>. }\footnotemark[2]\textsuperscript{1}},~~
 \textbf{Wei Ping\footnotemark[2]\textsuperscript{2}},~~
 \textbf{Peng Xu\textsuperscript{2}},~~
 \textbf{Zihan Liu\textsuperscript{2}}
   \vspace{.2cm}
 \\
  \textbf{Boxin Wang\textsuperscript{2}},~
 \textbf{Mohammad Shoeybi\textsuperscript{2}},~
 \textbf{Bo Li\textsuperscript{1}},~
 \textbf{Bryan Catanzaro\textsuperscript{2}}
\\
}

\begin{document}
\maketitle
\begin{abstract}
Long-context capabilities are essential for a wide range of applications, including document and video understanding, in-context learning, and inference-time scaling, all of which require models to process and reason over long sequences of text and multimodal data.
In this work, we introduce a efficient training recipe for building ultra-long context LLMs from aligned \emph{instruct} model, pushing the boundaries of context lengths from 128K to 1M, 2M, and 4M tokens.
Our approach leverages efficient continued pretraining strategies to extend the context window and employs effective instruction tuning to maintain the instruction-following and reasoning abilities.
Our \textbf{UltraLong-8B}, built on Llama-3.1-Instruct with our recipe, achieves state-of-the-art performance across a diverse set of long-context benchmarks.
Importantly, models trained with our approach maintain competitive performance on standard benchmarks, demonstrating balanced improvements for both long and short context tasks.
We further provide an in-depth analysis of key design choices, highlighting the impacts of scaling strategies and data composition. 
Our findings establish a robust framework for efficiently scaling context lengths while preserving general model capabilities.
We release all model weights at \url{https://ultralong.github.io/}.
\end{abstract}

\section{Introduction}


Large language models (LLMs) have demonstrated remarkable performance across a diverse range of text and multimodal tasks~\citep{hurst2024gpt,team2024gemini,liu2024deepseek,yang2024qwen2, dai2024nvlm}. However, many applications, such as document and video understanding~\citep{team2024gemini, azzolini2025cosmos}, in-context learning, and inference-time scaling~\citep{guo2025deepseek}, demand the ability to process and reason over extremely long sequences of tokens~\citep{team2024gemini, xu2023retrieval, yang2025qwen2,xu2024chatqa,lu2024controlled}. In these scenarios, the limited context window of LLMs poses a significant bottleneck, as critical information scattered across lengthy documents may be overlooked. This limitation motivates the need for models that can efficiently handle ultra-long contexts without sacrificing performance on standard tasks.

Recent trends in both industry and academia have focused on extending the context windows of LLMs. Proprietary systems like GPT-4o~\citep{hurst2024gpt} support context lengths of up to 128K tokens, while the reasoning model o1~\citep{jaech2024openai} further pushes this limit to 200K tokens to accommodate inference-time scaling. Other models, such as Claude 3.5 Sonnet~\citep{anthropic2024claude35} and Gemini 1.5 Pro~\citep{team2024gemini}, have demonstrated the feasibility of handling even larger contexts. Despite significant progress in open-access models, many efforts have been limited by the lack of detailed training data blends~\citep{dubey2024llama} and efficient extension recipes~\citep{gao2024train}. Moreover, evaluations of these models have often relied on synthetic benchmarks that do not fully capture their performance on real-world long-context tasks~\citep{gradientlongcontextllama3}.

\input{fig/pipe}

In this work, we present a systematic recipe for training ultra-long context language models. Our approach involves two key stages. The first stage, continued pretraining, extends the context window of LLMs to ultra-long lengths (up to 1M, 2M, and 4M tokens) by leveraging a specially curated corpus. We introduce techniques such as the use of special document separators during concatenation and apply YaRN-based~\citep{peng2023yarn} RoPE scaling to improve the model's ability to process long sequences. The second stage, instruction tuning, refines the model's instruction-following and reasoning capabilities using a high-quality, short-context supervised fine-tuning (SFT) dataset across general, mathematical, and coding domains.

We validate our approach using \basetext with 128K context window~\citep{dubey2024llama} as the starting point and conduct extensive evaluations on both synthetic and real-world long-context benchmarks, including RULER~\citep{hsieh2024ruler}, LV-Eval~\citep{yuan2024lv}, and InfiniteBench~\citep{zhang-etal-2024-bench}. Additionally, we benchmark our models on standard datasets such as MMLU~\citep{hendrycks2020measuring}, MATH~\citep{hendrycks2021measuring}, GSM-8K~\citep{cobbe2021gsm8k}, and HumanEval~\citep{chen2021evaluating} to ensure that the extended context does not compromise general task performance. Evaluation results demonstrate that our final models, \textbf{UltraLong-8B}, achieve state-of-the-art (SOTA) long-context performance while maintaining competitive performance on standard benchmarks.

Our contributions are summarized as follows:
\begin{itemize}
    \item We propose an efficient and scalable training recipe that extends the context window of LLMs to ultra-long lengths (up to 4M tokens) while preserving, and in some cases enhancing, performance on standard benchmarks.
    \item We introduce techniques such as the use of special document separators during data preparation and apply YaRN-based scaling for positional embeddings, both of which are shown through ablation studies to be essential for effective long-context modeling.
    \item We show that a one-step continued pretraining strategy is more efficient than a multi-step approach for context extension, consistently yielding superior results on both synthetic and real-world long-context benchmarks.
    \item We conduct extensive experiments on benchmarks including RULER, LV-Eval, InfiniteBench, MMLU, MMLU-Pro, MATH, GSM-8K, and HumanEval, showing that our UltraLong-8B models outperform existing baselines in both long-context and standard tasks.
\end{itemize}

The remainder of the paper is organized as follows. \Cref{sec:related_work} reviews related work on long-context language modeling and extension strategies. \Cref{sec:method} details our training methodology, including both the continued pretraining and instruction tuning stages. In \Cref{sec:baseline}, we describe the baseline models and evaluation benchmarks used in our experiments, while \Cref{sec:results} presents our experimental results. \Cref{sec:ablation} provides an in-depth analysis through ablation studies, and \Cref{sec:limitation,sec:conclusion} concludes the paper by discussing limitations and outlining directions for future research.

\section{Related Work}
\label{sec:related_work}

\subsection{Long-context methods}

Existing context extension strategies for long-context language models can be broadly categorized into three groups: exact attention methods, approximate attention methods, and approaches that incorporate additional modules.
Exact attention methods enhance the parameterization of the attention mechanism to support longer sequences. Techniques such as Position Interpolation (PI)~\citep{chen2023extending}, NTK-aware~\citep{bloc972023ntk}, Dynamic NTK~\citep{emozilla2023dynamically}, YaRN~\citep{peng2023yarn}, and CLEX~\citep{chen2023clex}—all based on RoPE~\citep{su2024roformer}—design position embeddings that enable length extension. These approaches can be applied either through fine-tuning or to frozen models.
In contrast, approximate attention methods adopt structured approximations to mitigate the computational cost of long-context processing. For example, LongLoRA~\citep{chen2023longlora} combines LoRA~\citep{hu2021lora} with Shifted Sparse Attention to reduce overhead, while LM-Infinite~\citep{han-etal-2024-lm} limits attention to a few tokens at the beginning of the text and a local window to remain within the pretrained length. Other approaches, such as Dual Chunk Attention~\citep{an2024training}, decompose attention into chunk-based modules to better capture the relative positional information, and some works~\citep{xu2023retrieval} leverage retrieval mechanisms to extract relevant blocks from long documents.
Additionally, methods that introduce extra modules~\citep{hwang2024transformerfam,ren2024samba} focus on compressing the information in the long input contexts.
In this work, we focus on exact attention techniques that accurately compute full attention over extended sequences, and we introduce an efficient training recipe to enable models to handle ultra-long contexts more effectively.


\subsection{Long-context LLMs}
Recent advancements in long-context LLMs include proprietary models such as GPT-4o~\citep{hurst2024gpt}, Gemini~\citep{team2024gemini}, and Claude~\citep{anthropic2024claude35}, which support extensive context windows and can process hundreds of thousands of tokens, though their closed-source nature limits reproducibility. Among open-source efforts, ProLong~\citep{gao2024train} employs NTK-aware scaling and trains on over 40B tokens, making it computationally expensive, while Gradient~\citep{gradientlongcontextllama3} uses a multi-step continued pretraining strategy that sacrifices standard task performance. 
ChatQA 2~\citep{xu2024chatqa} develops long-context LLMs that excel in both long-context understanding and retrieval-augmented generation.
In contrast, our work offers a balanced solution that extends the context window to ultra-long lengths while maintaining competitive performance on standard benchmarks through efficient continued pretraining and instruction tuning.

\section{Method}
\label{sec:method}

In this section, we present our training recipe for ultra-long context models, as illustrated in~\Cref{fig:pipe}. Our approach consists of two key stages: continued pretraining and instruction tuning. Based on \basetext~\citep{dubey2024llama}, the continued pretraining stage extends the context window of the model from 128K tokens to the target lengths (e.g., 1M, 2M, and 4M tokens). Subsequently, the instruction tuning stage refines the model to enhance its instruction-following and reasoning abilities. Together, these stages enable our models to effectively process ultra-long inputs while maintaining strong performance on both long and short-context tasks. 
More details can be found in~\Cref{sec:recipe}.

\subsection{Continued Pretraining for Context Length Extension}

In the first stage, we extend the context window of \basetext to the target length through continued pretraining.

\paragraph{Data Preparation and Document Concatenation.} 
We construct our long-context pretraining corpus following the methodology outlined by~\citet{fu2024data}. To emphasize long-context data, we downsample documents shorter than 4K tokens and upsample those longer than 8K tokens, resulting in a corpus of 1 billion tokens. These documents are then concatenated to form longer sequences corresponding to the target context lengths (e.g., 1M, 2M, and 4M tokens). During concatenation, we separate individual documents using special characters rather than the reserved beginning and ending tokens (\texttt{``<|begin\_of\_text|>''} and \texttt{``<|end\_of\_text|>''}). Furthermore, we do not apply the cross-document attention mask~\citep{gao2024train} during continued pretraining, allowing the model to attend to the entire input sequence. Our preliminary experiments indicate that this document separation strategy, combined with full attention, enables the model to adapt more effectively and efficiently to ultra-long contexts. Detailed ablation studies and analysis are provided in \Cref{sec:ablation_separator}.

\paragraph{RoPE Scaling.} To support ultra-long context lengths, we adopt a YaRN-based scaling approach~\citep{peng2023yarn} rather than the NTK-aware scaling strategies employed in previous work~\citep{xu2024chatqa,gao2024train,gradientlongcontextllama3}. We fix the hyperparameters as $\alpha=1$ and $\beta=4$, and compute the scale factor $s$ based on the target context length. We observed that the Llama-3.1 model's performance degrades when the input length approaches the maximum limit. To mitigate this, we employ a larger scaling factor for the RoPE embeddings, thereby better accommodating extended sequences.

\paragraph{Implementation Details.} 
We build long-context models targeting three context lengths: 1M, 2M, and 4M tokens. We set the RoPE scaling factors to $s=128$, $256$, and $512$ accordingly. Each model is trained on 1B tokens for one epoch using a learning rate of $3\times10^{-5}$.
For scalability, we train the models using the Megatron-LM framework~\citep{shoeybi2019megatron}.
To handle ultra-long input sequences, we adopt tensor parallelism~(TP) with $tp=8$ and leverage context parallelism~(CP) by setting $cp=4$ for the 1M model and $cp=16$ for the 2M and 4M models. 
Training is done on 256 NVIDIA H100 GPUs, with the 1M, 2M, and 4M models requiring approximately 5, 6, and 13 hours of training, respectively.

\subsection{Instruction Tuning}

In the second stage, we enhance the instruction-following and reasoning capabilities of our long-context language models through supervised fine-tuning (SFT)~\citep{ouyang2022training} on carefully curated  datasets.

\paragraph{Data Preparation.}  We subsample a high-quality blend of SFT datasets from \citep{liu2024acemath} by integrating and refining multiple open-source SFT datasets spanning three key domains: general, mathematics, and code.~\footnote{The full dataset is available for download at:\\ \url{https://huggingface.co/datasets/nvidia/AceMath-Instruct-Training-Data}. \\
We subsample our SFT data from the `general\_sft\_stage2' split.}
For the general domain, we incorporate data from ShareGPT~\citep{chiang2023vicuna,sharegptvicuna}, SlimOrca~\citep{SlimOrca,mukherjee2023orca}, EvolInstruct~\citep{xu2024wizardlm}, GPTeacher~\citep{gpteachergeneralinstruct}, AlpacaGPT4~\citep{peng2023instruction}, and UltraInteract~\citep{yuan2024advancing}.
In the math domain, our dataset includes OrcaMathWordProblems~\citep{mitra2024orca}, MathInstruct~\citep{yue2023mammoth}, and MetaMath~\citep{yu2023metamath}.
For the code domain, we incorporate Magicoder~\citep{wei2024magicoder}, WizardCoder~\citep{luo2023wizardcoder}, and GlaiveCodeAssistant~\citep{glaivecodeassistant}.

To further enhance the quality of our SFT dataset, we leverage OpenAI's GPT-4o~\citep{hurst2024gpt} and GPT-4o-mini~\citep{openai20244o} to refine the responses associated with these prompts. Finally, we perform rigorous data decontamination to ensure that our dataset does not include any prompts from benchmark test datasets.

Notably, our SFT blend exclusively comprises the short-context data described above, consisting of instances shorter than 8K tokens, without incorporating synthetic long-context instruction data as employed in~\citet{xu2024chatqa,zhao2024longskywork}, We find that relying solely on short-context data is sufficient to achieve strong results in our setting, aligning with observations from prior work~\citep{gao2024train}.

\paragraph{Implementation Details.}
We construct an SFT dataset comprising 100K examples. For every model extended to one of the three target context lengths, we use a batch size of 128 and a learning rate of $5\times10^{-6}$. Training is performed using the Megatron-LM framework~\citep{shoeybi2019megatron} on 256 NVIDIA H100 GPUs with tensor parallelism set to $tp=8$. Each training run completes in approximately 30 minutes.

\section{Baselines and Evaluation Benchmarks}
\label{sec:baseline}

\subsection{Long context models}
We compare our models against SOTA long context models built on the Llama family to ensure a fair and controlled evaluation of our training recipe.
\textbf{Llama-3.1} (\basetext)~\citep{dubey2024llama} serves as our base model and features a 128K context window. \textbf{ProLong} (\prolongtext)~\citep{gao2024train} is a long-context model built on Llama-3 with a 512K context window. This model is trained using two stages of continued pretraining and additional SFT, on a total of 41B tokens. \textbf{Gradient} (\gradienttext)~\citep{gradientlongcontextllama3} is another Llama-based long-context model, supporting a 1M context window. It is trained through four stages of continued pretraining on a total of 1.4B tokens, and additional SFT to strengthen its chat capabilities.
Focusing exclusively on models within the Llama family allows us to clearly isolate and demonstrate the effectiveness of our training recipe for extending context lengths, while ensuring that standard task performance remains competitive.

\input{fig/niah}

\subsection{Long context benchmarks}
We evaluate the long-context capabilities of our models using the following benchmarks.

\noindent \textbf{RULER}~\citep{hsieh2024ruler} is a benchmark that assesses long-context language models by generating synthetic examples with configurable sequence lengths across four task categories. The benchmark originally evaluates models within a 128K context window by computing the average accuracy across different input lengths. We adopt the same generation protocol and construct test cases with lengths up to 1M tokens, including 256K, 512K, and 1M. Since some of our baseline models support input lengths only up to 128K or 512K, we report the average accuracy for inputs shorter than 128K, 512K, and 1M tokens, respectively.

\noindent \textbf{LV-Eval}~\citep{yuan2024lv} is a long-context benchmark featuring five length levels up to 256K tokens, and focuses on two primary tasks: single-hop QA and multi-hop QA. We follow the evaluation protocol and compute the average F1 score across input lengths below 128K and 256K tokens.

\noindent \textbf{InfiniteBench}~\citep{zhang-etal-2024-bench} is a long-context benchmark with an average input length of around 200K tokens and a maximum length exceeding 2M tokens. It includes both synthetic and real-world tasks. As this benchmark does not provide scores for specific input length levels, we follow the standard evaluation protocol and compute the average score across all tasks.

\subsection{Standard benchmarks}
We assess the standard capabilities of our models across three domains. For the general domain, we evaluate using \textbf{MMLU}~\citep{hendrycks2020measuring} and \textbf{MMLU-Pro}~\citep{wang2024mmlu}, where we report the 5-shot accuracy. 
In the math domain, we employ two benchmarks: \textbf{MATH}~\citep{hendrycks2021measuring}, for which we report the 0-shot exact match accuracy, and \textbf{GSM-8K}~\citep{cobbe2021gsm8k}, for which we report the 8-shot exact match accuracy. For the code domain, we consider \textbf{HumanEval}~\citep{chen2021evaluating} and report the 0-shot pass@1 score.
We keep all the demonstrations the same when evaluating different models.

\section{Results}
\label{sec:results}

\input{tables/main_long}

In this section, we present the results and comparisons from extensive benchmark evaluations. We begin with the synthetic Needle in a Haystack (NIAH) test and then focus on both long-context and standard benchmarks.

\subsection{Needle in a Haystack}
We evaluate the long-context retrieval capabilities of our models using the Needle in a Haystack (NIAH) passkey retrieval test as defined in \citet{mohtashami2023landmark}.
This synthetic task serves as a threshold evaluation for assessing whether a language model can maintain awareness of critical information distributed throughout an extremely long input. In this task, the model is challenged to locate a simple passkey—such as a six-digit random number—embedded within a lengthy, nonsensical text sequence.

To quantify retrieval accuracy, we evaluate 40 different input sequence lengths. For each length, the passkey is randomly embedded at 10 uniformly distributed document depths. 
The results are shown in \Cref{fig:niah}. For our models, we evaluate input lengths up to 1M, 2M, and 4M tokens, while for baseline models, we only evaluate up to 1M tokens.
As illustrated in~\Cref{fig:niah_llama31,fig:niah_prolong,fig:niah_gradient}, among the baseline models, only \gradienttext passes the NIAH test, while \basetext and \prolongtext exhibit errors even within their claimed context lengths. In contrast, as shown in~\Cref{fig:niah_1m,fig:niah_2m,fig:niah_4m}, our UltraLong models achieve 100\% accuracy across all input lengths and depths, demonstrating robust long-context retrieval capability.


\subsection{Long context evaluation}

We present the evaluation results on RULER, LV-Eval, and InfiniteBench in \Cref{tab:long}. Bolded numbers indicate performance that exceeds all baseline models. Overall, our three models consistently achieve the highest scores in most cases. On the RULER benchmark, UltraLong models obtain the highest average scores for input lengths up to 512K and 1M tokens. For LV-Eval, our models yield the highest average F1 scores within both 128K and 256K token lengths. Additionally, we achieve the best performance on InfiniteBench.

These results confirm that our training recipe effectively extends the context window of language models to ultra-long inputs while maintaining similar performance on the original input length. Among the baseline models, Llama-3.1 is designed for a 128K input length and its performance degrades significantly when the input length exceeds 128K tokens. ProLong, built for a 512K context, performs worse than our models at 512K even though it is trained on substantially more tokens (41B tokens versus 1B tokens in our case). Gradient, the longest-context model among the baselines with support for a 1M input length, exhibits poor performance on LV-Eval and InfiniteBench, suggesting that its design may be overly tuned to synthetic tasks and compromises its effectiveness on real-world tasks. In contrast, our models consistently achieve higher scores across both synthetic (RULER) and hybrid (LV-Eval and InfiniteBench) benchmarks, underscoring the effectiveness and scalability of our approach.

\input{tables/main_standard}

\subsection{Standard capability evaluation}

We further evaluate our models on standard benchmarks across general, math, and code domains to ensure that extending the context length does not compromise short-context task performance. As shown in \Cref{tab:standard}, our models achieve performance that is comparable to or higher than that of the base model, \basetext, achieving average scores of 62.47, 61.06, and 60.95, respectively, compared to 61.45 for \basetext. Notably, our models demonstrate clear improvements on the MMLU and MATH benchmarks, while their performance on other benchmarks such as GSM-8K and HumanEval remains highly competitive.

In contrast, the baseline long-context models, Gradient and ProLong, experience considerable performance degradation on these standard tasks, with average scores of only 37.36 and 40.81, respectively. These results indicate that while our approach effectively extends the context window, it also preserves—and in some cases enhances—the model's general task capabilities. The significant drop in performance for \gradienttext and \prolongtext suggests that their methods for long-context training may come at the expense of general task performance.

Overall, our findings demonstrate that our training recipe successfully extends the context length of language models while maintaining strong performance on standard benchmarks.

\input{tables/continue_pretrain}

\input{tables/multi_stage}

\section{Ablation Studies}
\label{sec:ablation}

\paragraph{Special document separator helps efficient context extension.}
\label{sec:ablation_separator}
To assess the impact of using special characters as document separators during long-context continued pretraining, we conduct an ablation study in which we remove all document separators from our pretraining corpus while keeping all other settings unchanged. The model is trained on the same 1B-token corpus, and the results are shown in~\Cref{tab:ablation_continue}. We find that removing the special document separator leads to a consistent drop in performance across all benchmarks. For example, on the RULER benchmark within 1M token input length, the model without the separator scores 79.15, compared to 80.17 when the separator is used. Similar trends are observed in LV-Eval and InfiniteBench, where the absence of the special document separator results in performance degradation. These findings indicate that employing special characters as document separators—especially when enabling cross-document attention—enhances training efficiency, achieving higher performance at the same training cost.

\paragraph{YaRN-based scaling helps scalable context extension.}
\label{sec:ablation_yarn}
We also investigate the impact of scaling strategies on context extension by comparing our YaRN-based scaling with the NTK-aware scaling approach~\citep{xu2024chatqa,gao2024train,gradientlongcontextllama3}. In this experiment, we replace YaRN-based scaling with NTK-aware scaling using $\theta=3,580,165,449$, consistent with the configuration in Gradient. The model is again trained on the same 1B-token corpus, and the results are presented in~\Cref{tab:ablation_continue}. While NTK-aware scaling produces a slightly higher RULER score within 128K tokens (86.91 versus 85.63), it suffers a significant performance drop at extended lengths. For instance, on the RULER benchmark, the NTK-aware scaling variant achieves 76.62 within 1M tokens, compared to 80.17 with YaRN-based scaling. Similar declines are observed in LV-Eval and InfiniteBench. These results confirm that YaRN-based scaling offers a more robust and scalable method for extending the context window, enabling the model to maintain high performance even as the input length increases.

\paragraph{One-step continued pretraining is more effective for context extension.}
\label{sec:ablation_one_step}
We compare our one-step continued pretraining strategy against a multi-step approach~\citep{gao2024train,gradientlongcontextllama3} for extending the context window. Following the setting in~\citet{gao2024train}, we split the context extension into two continued pretraining stages. For the 1M model, we first extend the context to 512K tokens by training on 0.5B tokens, and then extend it to 1M tokens with an additional 0.5B tokens—resulting in a total of 1B tokens. We then compare this multi-step model with our one-step model that is directly extended to 1M tokens using 1B tokens of training data. Similarly, for the 2M model, we first extend to 1M and then to 2M, each stage using 0.5B tokens, and compare it with our one-step 2M model trained on 1B tokens. The results are shown in \Cref{tab:ablation_multi_stage}. We find that our one-step continued pretraining approach consistently outperforms the multi-step approach across all benchmarks. For example, our one-step 1M model achieves average RULER scores of 85.63, 82.28, and 80.17 within input lengths of 128K, 512K, and 1M, respectively, compared to 84.22, 79.83, and 77.52 for 1M model extended with 2 steps. Similarly, our one-step 2M model outperforms 2M model trained with 2 steps on both RULER and LV-Eval, with higher scores observed on InfiniteBench as well. These results indicate that a one-step extension strategy is more efficient and effective than a multi-step training process.

\section{Limitations and Future Work}
\label{sec:limitation}

Our current work focuses on SFT on instruction datasets during the instruction tuning stage, and does not explore reinforcement learning or preference optimization, which we leave for future work. Additionally, while we present an effective recipe for extending long-context capabilities, our work does not address safety alignment. As a result, potential risks such as the generation of harmful or misleading information and privacy concerns remain unmitigated. Future research will aim to integrate safety alignment mechanisms and explore advanced tuning strategies to further enhance both performance and trustworthiness.

\section{Conclusion}
\label{sec:conclusion}

In this work, we introduce an efficient and systematic training recipe for ultra-long context language models, extending context windows to 1M, 2M, and 4M tokens while maintaining competitive performance on standard benchmarks. Our approach combines efficient continued pretraining with instruction tuning to enhance both long-context understanding and instruction-following capabilities. Extensive experiments and ablation studies on our UltraLong-8B model series validate our design choices, including the use of special document separators, YaRN-based scaling, and an efficient one-step training strategy. We believe our framework sets a new standard for scalable long-context modeling and paves the way for future research aimed at improving long-context performance in practical applications.

\section*{Acknowledgments}

Chejian Xu and Bo Li are partially supported by the National Science Foundation under grant No. 1910100, No. 2046726, NSF AI Institute ACTION No. IIS-2229876, DARPA TIAMAT No. 80321, the National Aeronautics and Space Administration (NASA) under grant No. 80NSSC20M0229, ARL Grant W911NF-23-2-0137, Alfred P. Sloan Fellowship, the research grant from AI Safety Fund, Virtue AI, and Schmidt Science.


\bibliography{custom}

\newpage
\newpage
\newpage

\appendix
\onecolumn

\section{UltraLong Recipe}
\label{sec:recipe}

Table~\ref{tab:recipe} provides a detailed overview of our training recipe, which is divided into two stages: continued pretraining and instruction tuning. In the continued pretraining phase, we utilize per-source upsampled pretraining data as \citet{fu2024data} to extend the context window to 1M, 2M, and 4M tokens over 1B tokens of training. The model is initialized from \basetext with RoPE-based positional embeddings (base frequency \(5 \times 10^5\) and initial scaling factor \(s=8\)), and extended using scaling factors of \(s=128\), \(256\), and \(512\), with full attention applied without cross-document masking. In the instruction tuning stage, we fine-tune the model using a diverse dataset covering general, mathematics, and code domains, again training on 1B tokens with an Adam optimizer at a learning rate of \(5 \times 10^{-6}\).

\input{tables/recipe}

\end{document}

%% file: fig/pipe.tex
\begin{figure*}[t]
\includegraphics[width=0.8\linewidth]{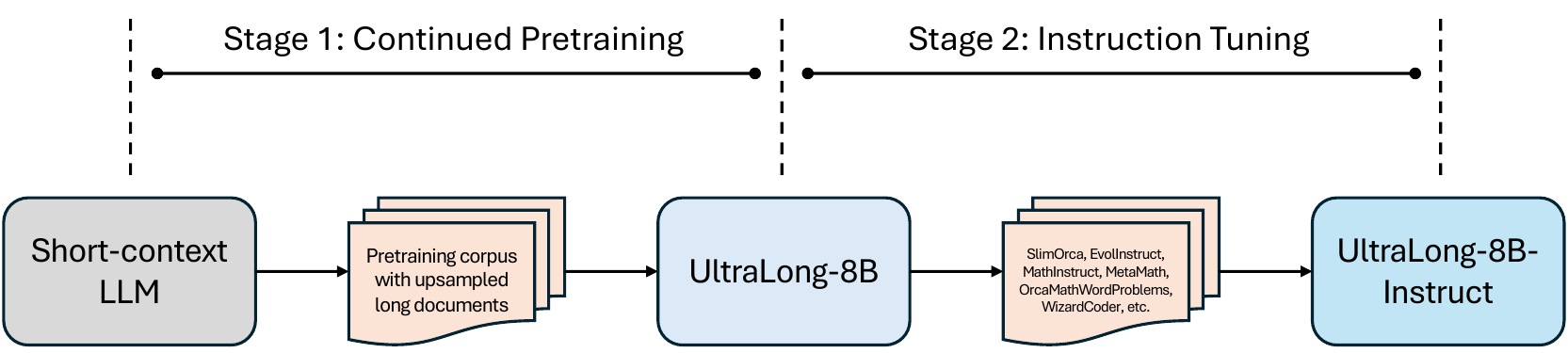}
\centering
\caption{Overview of our training pipeline. In Stage 1, the model's context window is extended through continued pretraining, leveraging techniques such as special document separators and YaRN-based scaling to handle ultra-long sequences. In Stage 2, instruction tuning is applied using a curated dataset to enhance the model's instruction-following and reasoning capabilities. This pipeline enables the development of language models that achieve good performance on both long-context and standard benchmarks.}
\label{fig:pipe}
\end{figure*}

%% file: fig/niah.tex





\begin{figure*}[t]
    \centering
    \begin{minipage}{0.49\textwidth}
        \centering
        \includegraphics[width=\linewidth]{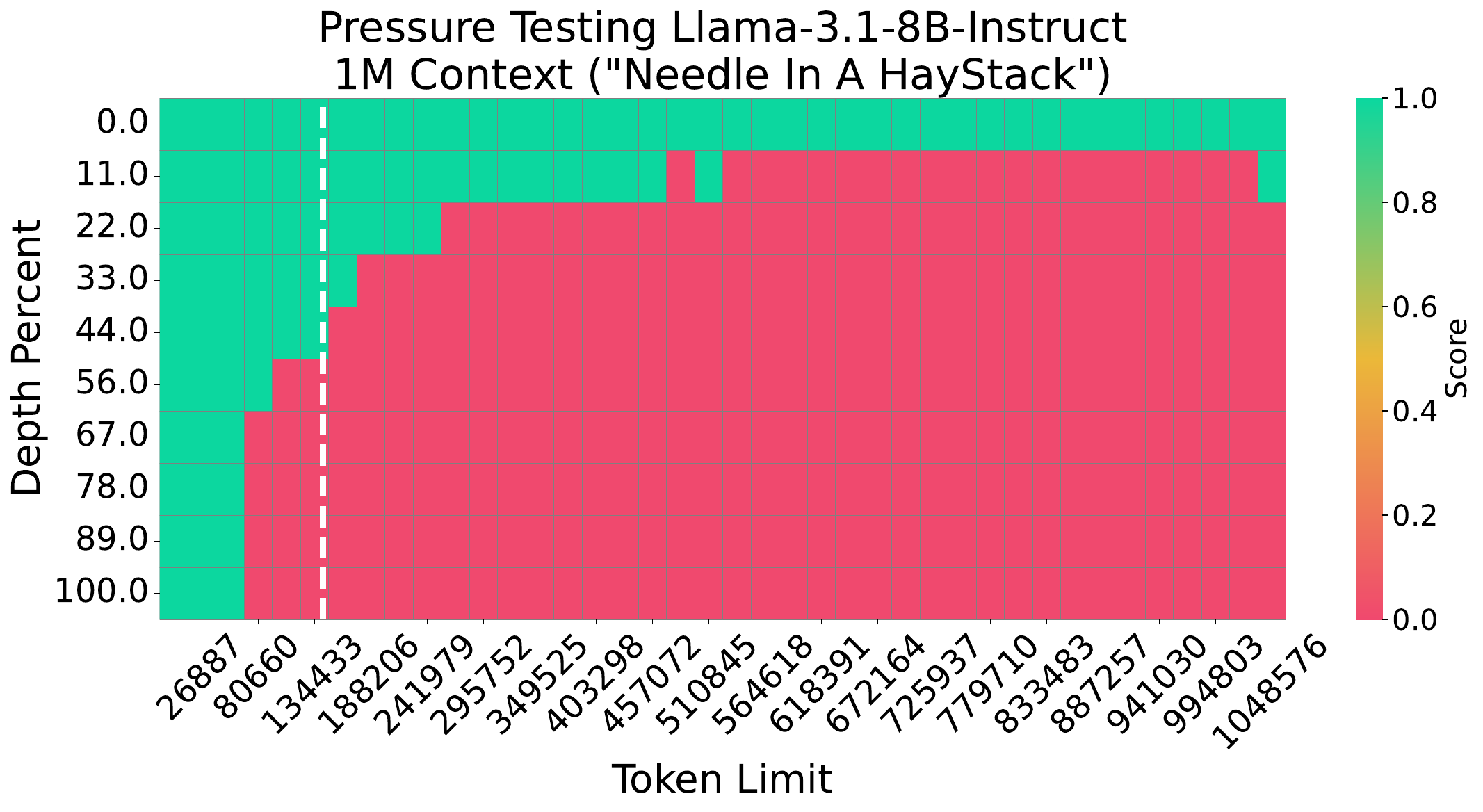}
        \subcaption{\base}
        \label{fig:niah_llama31}
    \end{minipage}
    \hfill
    \begin{minipage}{0.49\textwidth}
        \centering
        \includegraphics[width=\linewidth]{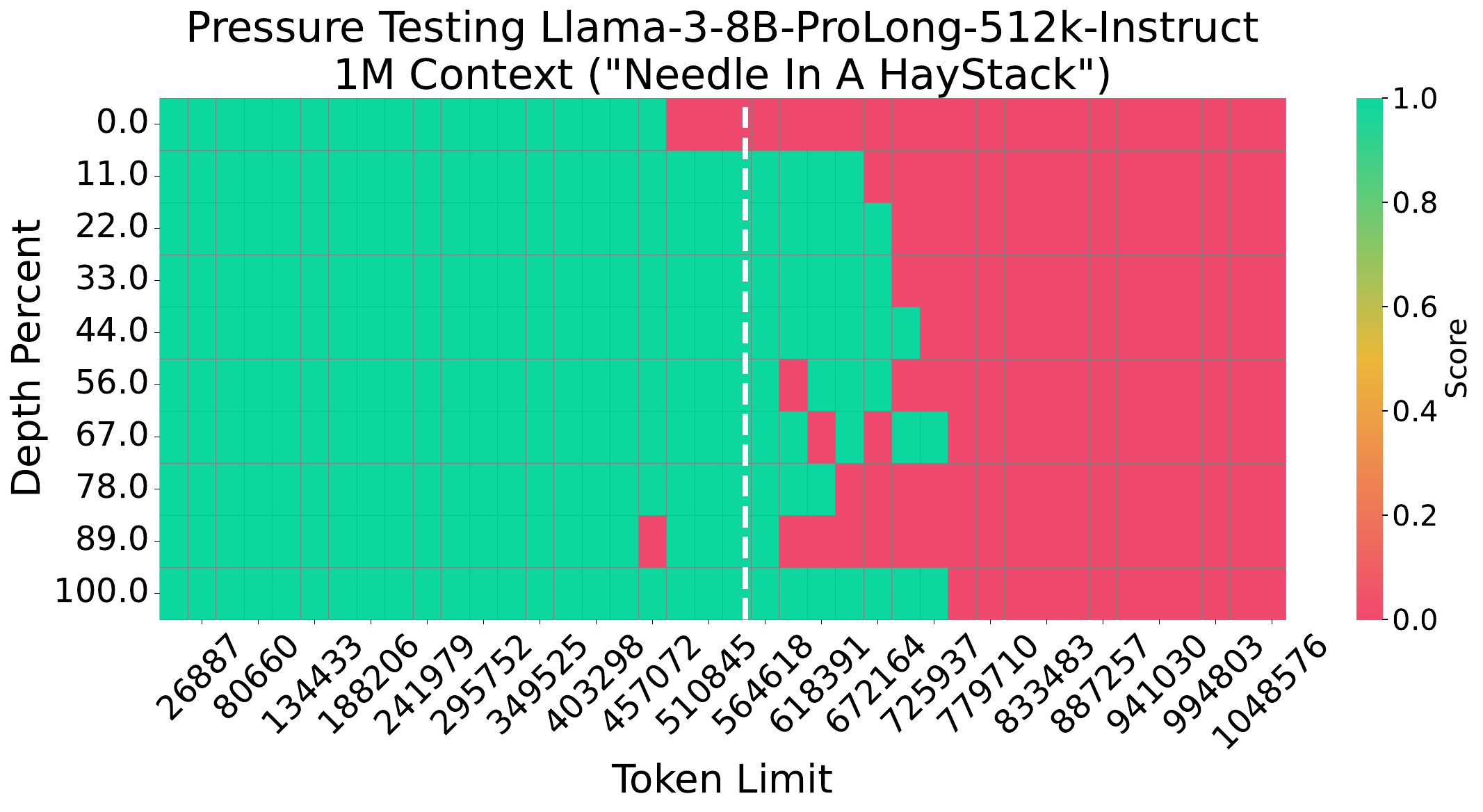}
        \subcaption{\prolong}
        \label{fig:niah_prolong}
    \end{minipage}

    \vspace{0.25cm}
    
    \begin{minipage}{0.49\textwidth}
        \centering
        \includegraphics[width=\linewidth]{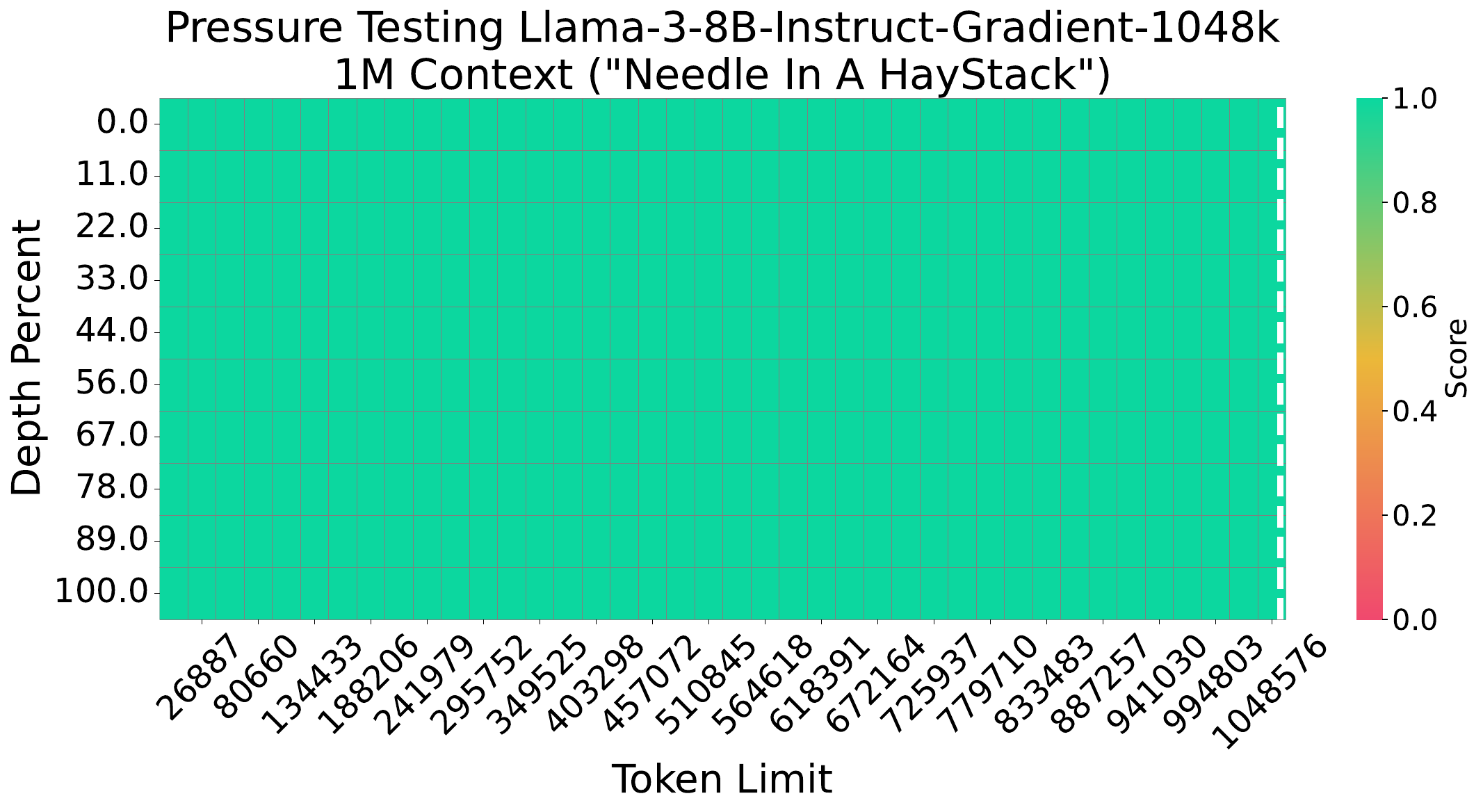}
        \subcaption{\gradient}
        \label{fig:niah_gradient}
    \end{minipage}
    \hfill
    \begin{minipage}{0.49\textwidth}
        \centering
        \includegraphics[width=\linewidth]{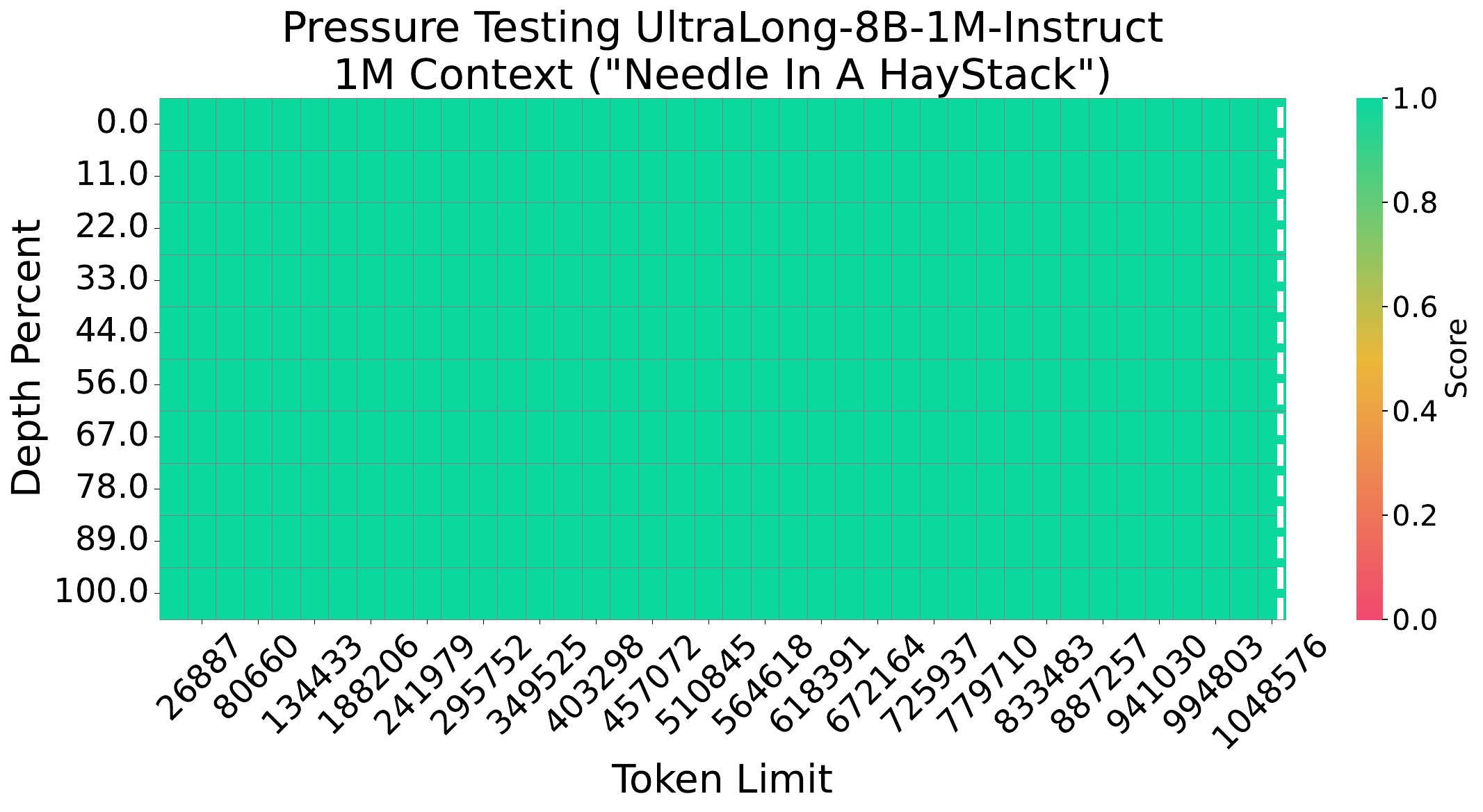}
        \subcaption{\modelI}
        \label{fig:niah_1m}
    \end{minipage}

    \vspace{0.25cm}
    
    \begin{minipage}{0.49\textwidth}
        \centering
        \includegraphics[width=\linewidth]{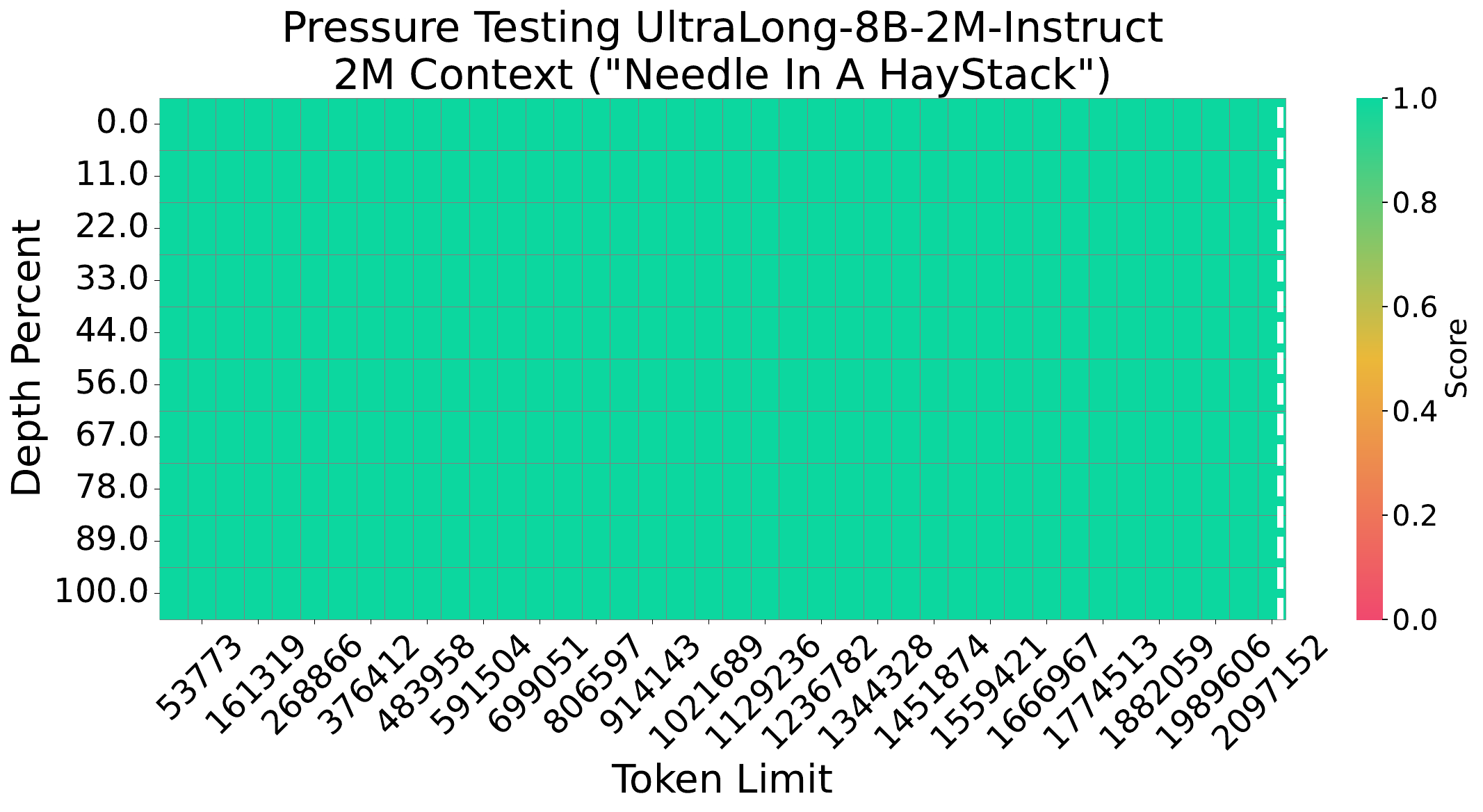}
        \subcaption{\modelII}
        \label{fig:niah_2m}
    \end{minipage}
    \hfill
    \begin{minipage}{0.49\textwidth}
        \centering
        \includegraphics[width=\linewidth]{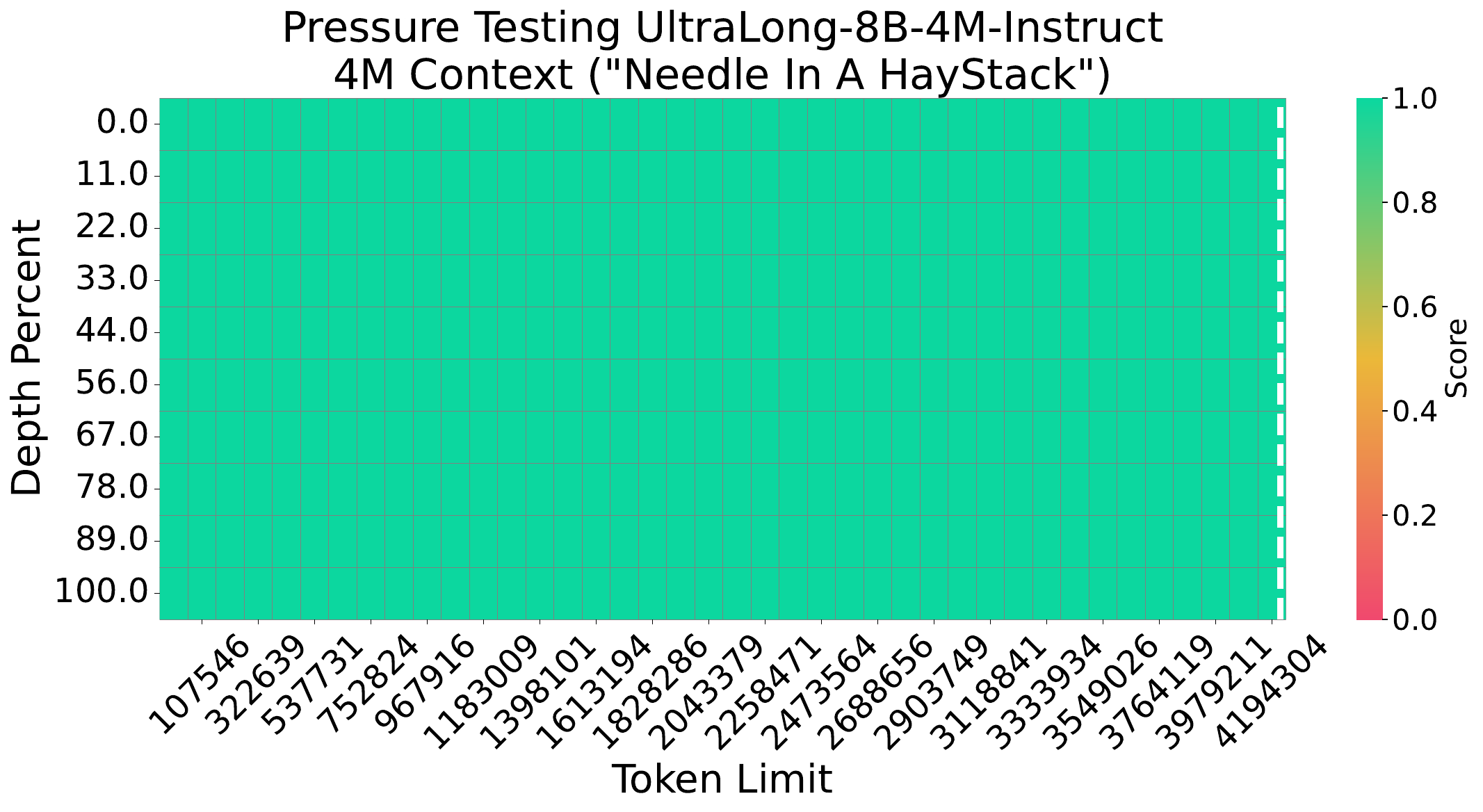}
        \subcaption{\modelIV}
        \label{fig:niah_4m}
    \end{minipage}

    \caption{Needle in a Haystack passkey retrieval test results. Three baseline models are evaluated up to a 1M token context, while our models are tested at their respective maximum context lengths of 1M, 2M, and 4M tokens. Our models achieve 100\% accuracy across all input lengths and depths, showing strong long-context retrieval capability.}
    \label{fig:niah}
\end{figure*}

%% file: tables/main_long.tex
\begin{table*}[t]
\centering
{
\resizebox{.95\textwidth}{!}{
    \begin{tabular}{lccccccccccc}
    \toprule
        \multirow{2}{*}{Model} & \multicolumn{3}{c}{RULER} & \multicolumn{2}{c}{LV-Eval} & \multirow{2}{*}{InfiniteBench} \\
        \cmidrule(lr){2-4} \cmidrule(lr){5-6}
        & < 128K & < 512K & < 1M & < 128K & < 256K & \\
        \midrule
        \base~(128K) & 88.3 & 68.2 & 61.3 & 26.55 & 23.32 & 24.66 \\
        \midrule
        \gradient & 82.4 & 79.8 & 77.8 & 14.14 & 13.48 & 28.60 \\
        \prolong & \textbf{88.9} & 79.6 & 71.2 & 25.59 & 24.65 & 23.54 \\
        \midrule
        \modelI & 86.6 & \textbf{81.6} & \textbf{79.1} & {28.07} & {27.02} & {32.14} \\
        \modelII & 85.0 & {80.2} & {78.2} & \textbf{30.03} & \textbf{28.90} & \textbf{32.49} \\
        \modelIV & 84.2 & {80.0} & {78.0} & {29.09} & {28.13} & {30.38} \\
        \bottomrule
    \end{tabular}
    }
}
\caption{Long context evaluation results on the RULER, LV-Eval, and InfiniteBench benchmarks. Average scores are reported within multiple input lengths—128K, 512K, and 1M for RULER, and 128K and 256K for LV-Eval—while a single aggregate score is provided for InfiniteBench. Bolded numbers indicate the highest performance among the models, demonstrating that our models consistently outperform the baselines.}
\label{tab:long}
\end{table*}

%% file: tables/main_standard.tex
\begin{table*}[t]
\centering
{
\setlength{\tabcolsep}{3.75pt}
\resizebox{0.95\textwidth}{!}{
    \begin{tabular}{lcccccc}
    \toprule
        Model & MMLU & MMLU-Pro  & MATH & GSM-8K & HumanEval & Avg \\
        \midrule
        \base~(128K) & 64.83 &  \textbf{44.33}  & 47.22 & \textbf{81.34} & \textbf{69.51} & 61.45 \\
        \midrule
        \gradient & 48.33 & 34.15 & 15.12 & 53.22 & 35.97 & 37.36 \\
        \prolong & 65.21 & 40.23 & 17.16 & 71.11 & 10.36 & 40.81 \\
        \midrule
        \modelI & {66.99} & {42.44} & \textbf{55.10} & {79.53} & {68.29} & \textbf{62.47} \\
        \modelII & \textbf{67.31} & {40.55}  & {51.36} & {79.00} & {67.07} & {61.06} \\
        \modelIV & 65.14 & {43.28} & {50.92} & {77.71} & {67.68} & {60.95} \\
        \bottomrule
    \end{tabular}
 }   
}
\caption{Evaluation results on standard benchmarks including MMLU, MMLU-Pro, MATH, GSM-8K, and HumanEval, along with the average score across these tasks. Our models demonstrate comparable or superior performance relative to the base model (\base), while the other long-context models exhibit significant degradation on standard tasks. Bolded numbers indicate the highest performance among the models.}
\label{tab:standard}
\end{table*}

%% file: tables/continue_pretrain.tex
\begin{table*}[t]
\centering
{
{
\resizebox{0.8\textwidth}{!}{
    \begin{tabular}{lcccccc}
    \toprule
        \multirow{2}{*}{Model} & \multicolumn{3}{c}{RULER} & \multicolumn{2}{c}{LV-Eval} & \multirow{2}{*}{InfiniteBench} \\
        \cmidrule(lr){2-4} \cmidrule(lr){5-6}
        & < 128K & < 512K & < 1M & < 128K & < 256K & \\
        \midrule
        \modelIbase & 85.63 & \textbf{82.28} & \textbf{80.17} & \textbf{27.60} & \textbf{26.40} & \textbf{26.25} \\
        \quad w/o special separator  & 85.47 & 81.63 & 79.15 & 26.06 & 24.85 & 22.75 \\
        \quad w/ NTK-aware scaling & \textbf{86.91} & 80.27 & 76.62 & 22.34 & 21.24 & 20.18 \\
        \bottomrule
    \end{tabular}
    }
  }
}
\caption{Ablation study results on our continued pretraining strategy for long-context extension. We compare our configuration with two ablated variants: one without the special document separator and one using NTK-aware scaling instead of the YaRN-based scaling. Performance is evaluated on the RULER benchmark for input lengths below 128K, 512K, and 1M tokens, on LV-Eval for input lengths below 128K and 256K tokens, and on InfiniteBench. Bolded scores indicate the best performance among the variants. These results demonstrate the importance of both the special document separator and YaRN-based scaling in achieving robust long-context performance.}
\label{tab:ablation_continue}
\end{table*}

%% file: tables/multi_stage.tex
\begin{table*}[t]
\centering
{
{
\resizebox{0.8\textwidth}{!}{
    \begin{tabular}{lcccccc}
    \toprule
        \multirow{2}{*}{Model} & \multicolumn{3}{c}{RULER} & \multicolumn{2}{c}{LV-Eval} & \multirow{2}{*}{InfiniteBench} \\
        \cmidrule(lr){2-4} \cmidrule(lr){5-6}
        & < 128K & < 512K & < 1M & < 128K & < 256K & \\
        \midrule
        \modelIbasemulti & 84.22 & 79.83 & 77.52 & 25.15 & 23.97 & \textbf{26.81} \\
        \modelIbase & \textbf{85.63} & \textbf{82.28} & \textbf{80.17} & \textbf{27.60} & \textbf{26.40} & 26.25 \\
        \midrule
        \modelIIbasemulti & 84.60 & 80.12 & 78.18 & 25.68 & 24.72 & 24.91 \\
        \modelIIbase & \textbf{86.30} & \textbf{82.75} & \textbf{80.8} & \textbf{26.00} & \textbf{24.86} & \textbf{25.22} \\
        \bottomrule
    \end{tabular}
    }
  }
}
\caption{Comparison of multi-step and one-step continued pretraining strategies for context extension. Results are presented on the RULER benchmark (for inputs below 128K, 512K, and 1M tokens), LV-Eval (for inputs below 128K and 256K tokens), and InfiniteBench. Bolded scores indicate superior performance within each model pair. With the same training cost, we find that the one-step approach consistently outperforms the multi-step variants.}
\label{tab:ablation_multi_stage}
\end{table*}

%% file: tables/recipe.tex
\begin{table*}[t]
\centering
{
{
\resizebox{\textwidth}{!}{
    \begin{tabular}{lll}
    \toprule
        \multicolumn{3}{c}{\textbf{Continued Pretraining}} \\
        \midrule
        \textbf{Data} & \multicolumn{2}{l}{Per-source upsampled pretraining corpus. Documents are separated using \texttt{``<s>''}.} \\
        \textbf{Length} & \multicolumn{2}{l}{1M, 2M, and 4M} \\
        \textbf{Steps} & \multicolumn{2}{l}{1B tokens} \\
        \textbf{Model} & Initialization: & \basetext (RoPE base freq. \(5 \times 10^5\), RoPE scaling factor $s=8$) \\
        & RoPE scaling: & $s=128$, $256$, and $512$ \\
        & Attention: & Full attention without cross-document attention masking \\
        \textbf{Optim.} & \multicolumn{2}{l}{Adam ($\beta_1 = 0.9$, $\beta_2 = 0.95$)} \\
        & LR: & 3e-05 \\
        \midrule
        \multicolumn{3}{c}{\textbf{Instruction Tuning}} \\
        \midrule
        \textbf{Data} & General: & ShareGPT, SlimOrca, EvolInstruct, GPTeacher, AlpacaGPT4, and UltraInteract \\
        & Mathematics: & OrcaMathWordProblems, MathInstruct, and MetaMath \\
        & Code: & Magicoder, WizardCoder, and GlaiveCodeAssistant \\
        \textbf{Steps} & \multicolumn{2}{l}{1B tokens} \\
        \textbf{Optim.} & \multicolumn{2}{l}{Adam ($\beta_1 = 0.9$, $\beta_2 = 0.95$)} \\
        & LR: & 5e-06 \\
        \bottomrule
    \end{tabular}
    }
  }
}
\caption{Overview of our training recipe for UltraLong-8B-Instruct models.}
\label{tab:recipe}
\end{table*}